\documentclass[10pt,twocolumn,letterpaper]{article}

\usepackage{wacv}              

\usepackage{graphicx}
\usepackage{amsmath}
\usepackage{amssymb}
\usepackage{booktabs}

\usepackage[pagebackref,breaklinks,colorlinks]{hyperref}

\usepackage[capitalize]{cleveref}
\crefname{section}{Sec.}{Secs.}
\Crefname{section}{Section}{Sections}
\Crefname{table}{Table}{Tables}
\crefname{table}{Tab.}{Tabs.}

\def\wacvPaperID{1545} 
\def\confName{WACV}
\def\confYear{2024}

\begin{document}

\title{Neural Image Compression Using Masked Sparse Visual Representation}

\author{Wei Jiang\\
Futurewei Technologies Inc.\\
Santa Clara, CA\\
{\tt\small wjiang@futurewei.com}
\and
Wei Wang\\
Futurewei Technologies Inc.\\
Santa Clara, CA\\
{\tt\small rickweiwang@futurewei.com}
\and
Yue Chen\\
Futurewei Technologies Inc.\\
Santa Clara, CA\\
{\tt\small ychen@futurewei.com}
}
\maketitle

\begin{abstract}
We study neural image compression based on the Sparse Visual Representation (SVR), where images are embedded into a discrete latent space spanned by learned visual codebooks. By sharing codebooks with the decoder, the encoder transfers integer codeword indices that are efficient and cross-platform robust, and the decoder retrieves the embedded latent feature using the indices for reconstruction. Previous SVR-based compression lacks effective mechanism for rate-distortion tradeoffs, where one can only pursue either high reconstruction quality or low transmission bitrate. We propose a Masked Adaptive Codebook learning (M-AdaCode) method that applies masks to the latent feature subspace to balance bitrate and reconstruction quality. A set of semantic-class-dependent basis codebooks are learned, which are weighted combined to generate a rich latent feature for high-quality reconstruction. The combining weights are adaptively derived from each input image, providing fidelity information with additional transmission costs. By masking out unimportant weights in the encoder and recovering them in the decoder, we can trade off reconstruction quality for transmission bits, and the masking rate controls the balance between bitrate and distortion. Experiments over the standard JPEG-AI dataset demonstrate the effectiveness of our M-AdaCode approach.
\end{abstract}

\section{Introduction}
\label{sec:intro}

\begin{figure*}
  \centering
  \begin{subfigure}{\linewidth}
   \includegraphics[width=\linewidth]{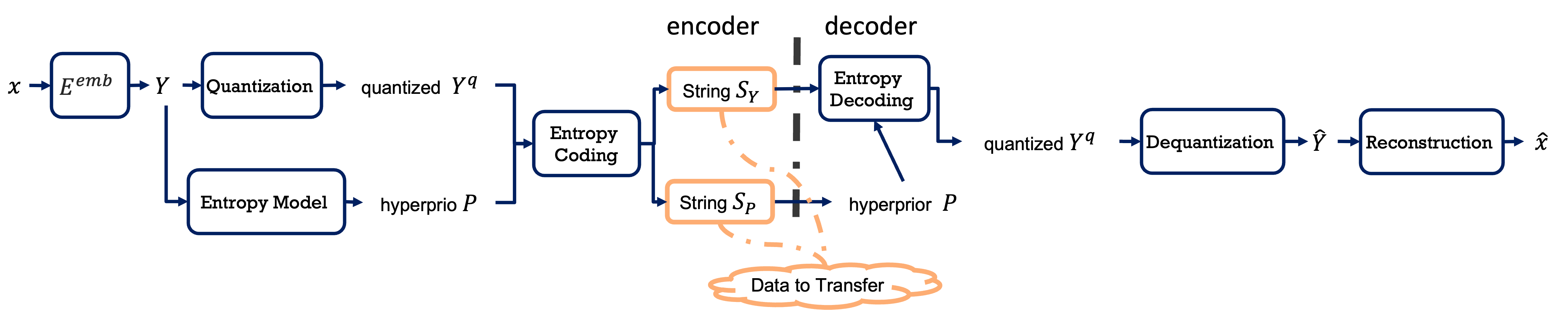}      
   \caption{The hyperprior framework requires deterministic and strictly consistent calculation of hyperprior $P$ in encoder and decoder.}\label{fig:hyperprior_architecture}  \vspace{.5em}
  \end{subfigure}
  \begin{subfigure}{\linewidth}
   \includegraphics[width=\linewidth]{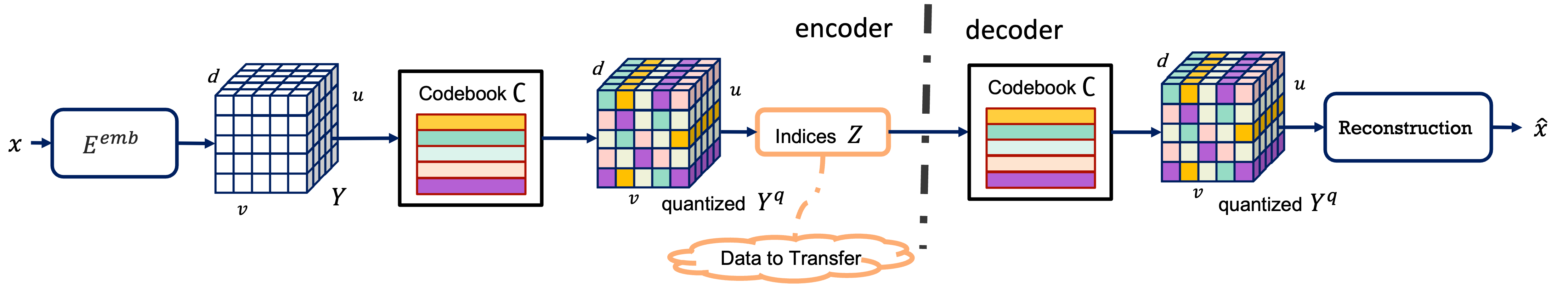}      
   \caption{SVR-based compression using one semantic-class-agnostic codebook \cite{MAGE} has low bitrate and low reconstruction quality.}\label{fig:vqgan_architecture}  \vspace{.5em}
  \end{subfigure}
  \begin{subfigure}{\linewidth}
   \includegraphics[width=\linewidth]{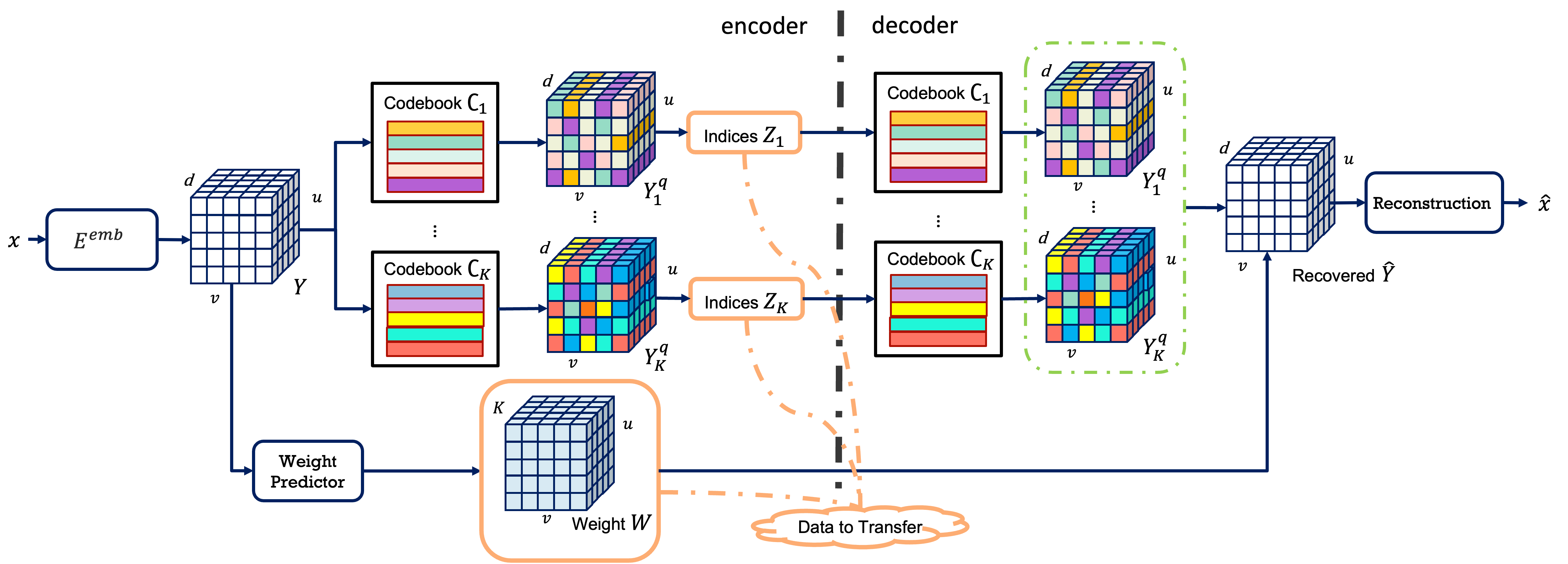}      
   \caption{SVR-based compression using semantic-class-dependent codebooks and image-adaptive weights \cite{AdaCode} has high reconstruction quality and high bitrate.}\label{fig:adacode_architecture}\vspace{.5em}
    \end{subfigure}
  \begin{subfigure}{\linewidth}
   \includegraphics[width=\linewidth]{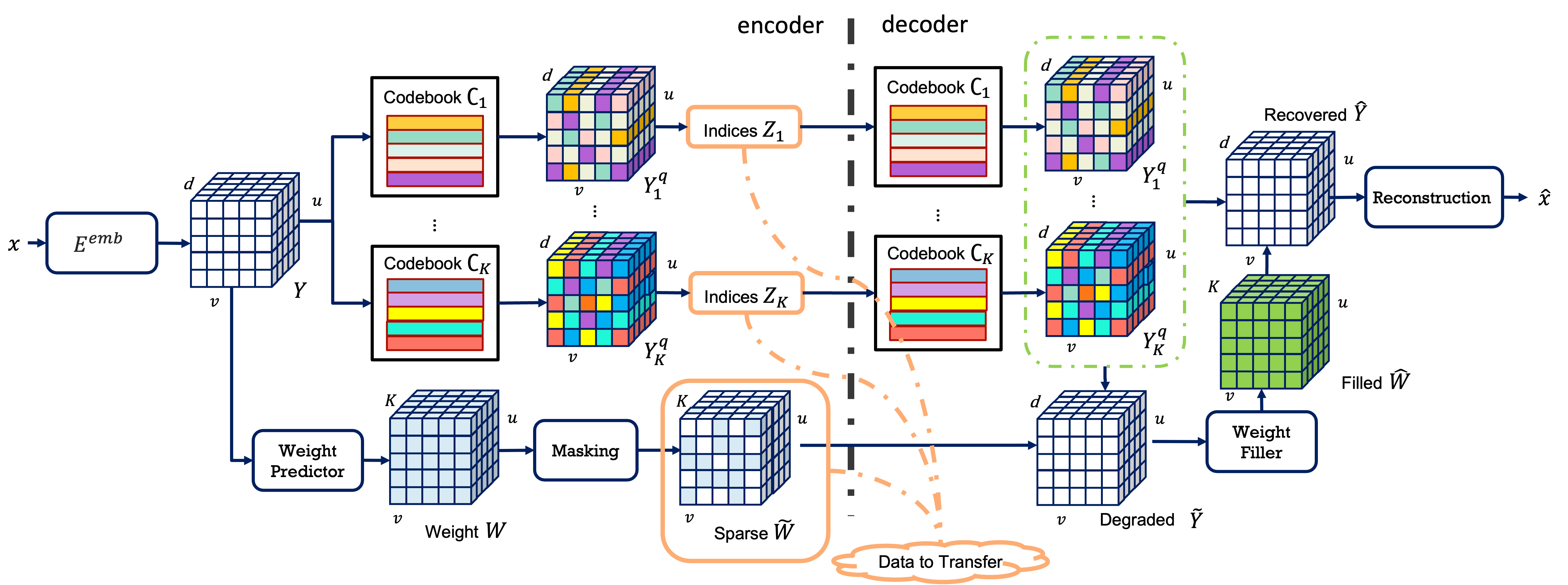}      
   \subcaption{SVR-based compression using M-AdaCode with flexible weight masking and refilling for tradeoffs between bitrate and reconstruction quality.}\label{fig:our_architecture}
      \end{subfigure}
   \caption{Different neural image compression frameworks.}
   \label{fig:architecture}
\end{figure*}

Neural image compression (NIC) has been actively studied in recent years. Using neural networks (NN), the encoder transforms the input image into a compact latent representation, based on which the decoder reconstructs the output image. NIC has two general research topics: (1) how to learn an effective and expressive latent representation, and (2) how to quantize and encode the latent representation for efficient transmission. So far, the most popular framework is based on hyperpriors \cite{hyperprior} (shown in Figure~\ref{fig:hyperprior_architecture}). An entropy model is used to encode/decode the quantized latent, which marries classical entropy coding with NN-based representation learning in a Variational AutoEncoder (VAE) structure. Many improvements have been made to the entropy model \cite{checkerboard,hypertransformer,VCT2022} to speedup computation and improve reconstruction quality.

In this work, we investigate a different framework for NIC based on the Sparse Visual Representation (SVR) (shown in Figure~\ref{fig:our_architecture}). We learn discrete generative priors as visual codebooks, and embed images into a discrete latent space spanned by the codebooks. By sharing the learned codebooks between the encoder and decoder, images can be mapped to integer codeword indices in the encoder, and the decoder can use these indices to retrieve the corresponding codeword latent feature for reconstruction. 

One major benefit of the SVR-based compression is the robustness to heterogeneous platforms by transferring integer indices. One caveat of the hyperprior framework is the extreme sensitivity to small differences between the encoder and decoder in calculating the hyperpriors $P$ \cite{entropyerrorICLR2019}. Even perturbations caused by floating round-off error can lead to catastrophic error propagation in the decoded latent feature $\hat{Y}$.  Most works simply assume homogeneous platforms and deterministic CPU calculation in the entropy model, which is unfortunately impractical. In real applications, senders and receivers usually use different hardware or software platforms where the numerical round-off difference well exists, and not using GPU to avoid the non-deterministic GPU calculation largely limits the computation speed. Only a few works have addressed this problem, \eg, by using integer NN to prevent non-deterministic GPU computation \cite{entropyerrorICLR2019} or by designing special NN modules that are friendly to CPU computation to speed up inference \cite{cpufriendly}. However, such solutions cannot be flexibly generalized to arbitrary network architectures. In comparison, SVR-based compression not only avoids the computational sensitive entropy model, but also brings additional benefits from SVR-based restoration, such as the improved robustness against input image degradations, and the freedom of expanding latent feature dimensions without increasing bitrates. 

In particular, we address the challenging dilemma of previous SVR-based compression in trading off bitrate and distortion: it is difficulty to achieve high-quality (HQ) reconstruction using one low-bitrate semantic-class-agnostic codebook, and it is difficulty to achieve low bitrate using multiple HQ semantic-class-dependent codebooks. Due to the complexity of visual content in natural images, the expressiveness and richness of one semantic-class-agnostic codebook (\eg, MAsked Generative Encoder as MAGE \cite{MAGE}) limits the reconstruction quality, while the additional image-adaptive information for recovering a rich feature for HQ reconstruction (\eg, image-Adaptive Codebook learning as AdaCode \cite{AdaCode}) consumes too many bits to transfer. 

We propose a Masked Adaptive Codebook learning (M-AdaCode) method for practical SVR-based compression, which applies masks to the latent feature subspaces to balance bitrates and reconstruction quality. Specifically, we build our method on top of AdaCode \cite{AdaCode} by adding an effective weight masking and refilling mechanism. A set of semantic-class-dependent basis codebooks are learned, and a weight map to combine these basis codebooks are adaptively determined for each input image. Adaptively combing the rich codebooks provides additional fidelity information for HQ reconstruction, but with high bit costs due to the transmission overhead of the dense weight map. By masking out unimportant weights in the encoder and recovering the weight map later in the decoder, we can reduce the transmission bits by compromising reconstruction performance. The masking rate controls the tradeoff between bitrate and reconstruction distortion. As shown in Figure~\ref{fig:svr_performance}, our method practically operates over a variety of bitrates, in contrast to previous SVR-based compression that only works in ultra-low or high bitrate ranges. 

Our M-AdaCode can also be seen as a method of Masked Image Modeling (MIM) \cite{MAE,MAGE}. Instead of applying masks in the spatial domain, we apply masks over latent feature subspaces. Using the redundant information in the latent space the HQ feature can be recovered from the degraded masked version, so that the masked SVR has improved representation efficiency to reduce transmission costs. We evaluate our approach over the standard JPEG-AI dataset \cite{JPEG-AI-data}. Our method is compared with the State-Of-The-Art (SOTA) class-agnostic SVR method MAGE \cite{MAGE} that uses spatial-masking MIM, and with the SOTA class-dependent SVR method AdaCode \cite{AdaCode} that uses a dense weight map. Experiments demonstrate the effectiveness of our M-AdaCode method.

\section{Related Works}

\subsection{Sparse Visual Representation Learning}
Discrete generative priors have shown impressive performance in image restoration tasks like  super-resolution \cite{FeMaSR}, denoising \cite{VQGAN}, compression \cite{JiangNTIRE23} \textit{etc}. By embedding images into a discrete latent space spanned by learned visual codebooks, the SVR has improved robustness to various image degradations. For instance, VQ-VAE \cite{VQVAE} learns a highly compressed codebook by a vector-quantized autoencoder. VQGAN \cite{VQGAN} further improves restoration quality by using Generative Adversarial Networks (GAN) with adversarial and perceptual loss. In general, it is difficult to learn a single general codebook for all image categories. Natural images have very complicated visual content, and a class-agnostic codebook usually has limited representation power for HQ reconstruction. Therefore, most methods focus on specific image categories (\eg, faces, architectures). For instance, SVR has achieved great success in face generation due to the highly structured characteristics of human faces, where an HQ codebook can be learned with generic and rich details for HQ face restoration \cite{RestoreFormer,CodeFormer2022}. 

For general natural images, to improve the restoration power of SVR, the recent AdaCode method \cite{AdaCode} uses an image-adaptive codebook learning approach. Instead of learning a single codebook for all categories of images, a set of basis codebooks are learned, each corresponding to a semantic partition of the latent space. A weight map to combine such basis codebooks are adaptively determined for each input image. By learning the semantic-class-guided codebooks, the semantic-class-agnostic restoration performance can be largely improved.

\subsection{Neural Image Compression}

There are two main research topics for NIC: how to learned an image latent representation, and how to quantize and encode the latent representation. One most popular framework is based on hyperpriors \cite{hyperprior}, where the image is transformed into a dense latent representation, and an entropy model encodes/decodes the quantized latent representation for efficient transmission. Many improvements have been made to improve the transformation for computing the latent \cite{windowtransformer,cheng2020,VCT2022} and/or the entropy model \cite{checkerboard,hypertransformer,VCT2022}. GAN has also been used for learning a good transformation \cite{GANLIC19,GANLIC20,RateDistortionPerception}. However, studies show that there are complex competing relations among bitrate, distortion, and perceptual quality \cite{PerceptionDistortion,RateDistortionPerception}. As a result, previous GAN-based NIC methods focus on very low-bitrate scenarios where low fidelity is less important than the good perceptual quality from generated textures and details.

One vital issue of the hyperprior framework is the extreme sensitivity to small differences between the encoder and decoder in calculating the hyperpriors \cite{entropyerrorICLR2019}. Even floating round-off error can lead to catastrophic error propagation in the decoded latent feature. The problem is largely overlooked, where most works simply assume homogeneous platforms and deterministic CPU calculation. Some work uses integer NN to prevent non-deterministic GPU computation \cite{entropyerrorICLR2019}. Some work designs special NN module that is computational friendly to CPU to speed up inference \cite{cpufriendly}. However, such solutions cannot be easily generalized to arbitrary network architectures. 

\subsection{SVR-based Compression}

SVR is intuitively suitable for compression among GAN-based generative methods. SVR represents images by codeword indices, based on which the decoder can retrieve the corresponding codeword feature for reconstruction. The integer indices are easy to transfer, and are robust to small computation differences in heterogeneous hardware and software platforms.  

However, due to the difficulty of learning SVR for HQ restoration over general images, previous methods use SVR for very low-bitrate cases, where reconstruction with low fidelity yet good perceptual-quality is tolerated. For example, MIM is combined with product quantization of VQ-VAE in \cite{MIMLIC} to achieve extreme compression rates. Other methods focus on special content categories that can be better modeled by SVR, such as human faces. For example, face reenactment is used to compress face videos  based on codebooks of facial keypoints \cite{Oneshot2021}. CodeFormer face restoration \cite{CodeFormer2022} is used to combine a VQGAN with highly compressed low-quality features to trade off perceptual quality and fidelity  \cite{JiangNTIRE23}. 

As for general images, to the best of our knowledge, no existing work studies SVR-based compression with normal bitrates. Although the AdaCode method \cite{AdaCode} can achieve high restoration quality, it is not compression-friendly due to the high transmission overhead for the predicted image-adaptive weight map. 

\subsection{Masked Image Modeling}

MIM has been shown effective in learning HQ visual representations via self-supervised learning. Early methods like MAE \cite{MAE} and CMAE \cite{CMAE} favor the performance of the representations on downstream tasks instead of the quality of the reconstructed images. 
The recent MAGE \cite{MAGE} learns a generic VQGAN representation by a single token-based MIM framework with variable masking ratios, which improves unconditioned image generation performance.

\section{Approach}

The general architecture of the baseline SVR-based image compression framework can be summarized in Figure~\ref{fig:vqgan_architecture}. An input image $X\!\in\!\mathbb{R}^{w\times h\times c}$ is first embedded into a latent feature $Y\!\in\!\mathbb{R}^{u \times v\times d}$ by an embedding network $E^{emb}$. Using a learned codebook $\mathcal{C}\!=\!\{c_l\!\in\!\mathbb{R}^d\}$, the latent $Y$ is further mapped into a discrete quantized latent feature $Y^q\!\in\!\mathbb{R}^{u \times v\times d}$. Specifically, each super-pixel $y^q(l)$ ($l\!=\!1,\ldots,u\times v$) in $Y^q$ corresponds to a codeword $c_l\!\in\!\mathcal{C}$ that is closest to the corresponding latent feature $y(l)$ in $Y$: 
\begin{equation}
    c_l= \mathit{argmin}_{c_i\in\mathcal{C}} D(c_i,y(l))). \nonumber
\end{equation}
Since $y^q(l)$ can be represented by the index $z_l$ of the codeword $c_l$, the entire $Y^q$ can be mapped to an $n$-dim vector $Z$ of integers, $n\!=\!u\times v$. $Z$ can be efficiently transmitted to the decoder with very little bit consumption, \eg, 10 bits/super-pixel for a codebook with 1024 codewords, and the compression rate can be quite high. On the decoder side, using the codebook $\mathcal{C}$, the quantized feature $Y^q$ is first retrieved based on the received codeword indices $Z$, and then a reconstruction network reconstructs the output image $\hat{x}$ based on $Y^q$. One example of this baseline SVR-based compression method is MAGE \cite{MAGE}, which uses MIM to learn a general SOTA visual codebook for general image reconstruction with very low bitrates. 


Aiming at improving the quality of the learned SVR for general image restoration, the AdaCode method  \cite{AdaCode} (as described in  Figure~\ref{fig:adacode_architecture}) learns a set of basis codebooks $\mathcal{C}_1,
\ldots,\mathcal{C}_{K}$, each corresponding to a semantic partition of the latent space. For each individual input, a weight map $W\!\in\!\mathbb{R}^{u\times v\times K}$ is computed to combine the basis codebooks for adaptive image restoration. Specifically, the embedded latent feature $Y$ is mapped to a set of quantized latent features $Y^q_1,\ldots,Y^q_K$ using each of the basis codebooks, respectively. Then a recovered latent $\hat{Y}$ is computed as a reconstructed version of latent $Y$, where for each super-pixel $\hat{y}(l)$  in the recovered $\hat{Y}$ ($l\!=\!1,\ldots,u\times v$):
\begin{equation}
    y(l)= \sum\nolimits_{j=1}^{K} w_{j}(l)y^q_{j}(l), \label{Eqn:weightedcombine}
\end{equation}
where $w_j(l)$ is the weight of the $j$-th codebook for the $l$-th super-pixel in $W$.  

This framework generates a more expressive recovered latent $\hat{Y}$ that preserves the fidelity cue of each input image than using a single semantic-class-agnostic codebook, and achieves SOTA reconstruction performance. However, it is not suitable for compression. The weight map $W$ needs to be transmitted for each input image, which consumes too many bits. As a result, AdaCode operates in the very high-bitrate range when used for compression. 

We propose a practical SVR-based compression framework that can operate in normal bitrate range. The main target is to recover a rich latent $\hat{Y}$ on the decoder side with as little transmitted data as possible. This is in comparison to the extreme case of MAGE that does not use any information to recover a rich latent, or Adacode that uses a dense weight map but ignores transmission costs.  Figure~\ref{fig:our_architecture} gives the detailed architecture of our M-AdaCode method. We use a weight masking and refilling mechanism. The encoder masks out unimportant weights in the weight map to reduce the amount of bits to transfer, which results in a degraded latent $\Tilde{Y}$ on the decoder side. Then the decoder re-predicts a full weight map $\hat{W}$ based on the degraded $\Tilde{Y}$ for combining codebooks, and computes the recovered latent $\hat{Y}$ for final image reconstruction. The masking rate controls the bitrate, ranging from using full weight map as AdaCode to only one codebook similar to MAGE. 

From another perspective, our M-AdaCode can be seen as an MIM method. Instead of applying masks in the spatial domain, we apply masks over latent feature subspaces, and use the redundant information in the feature subspace to recover the HQ latent feature from the degraded masked version. By controlling the masking rate, we tune the representation efficiency of SVR by trading off reconstruction quality for transmission bits.

\subsection{Weight Masking and Refilling}

Let $m$ denote the number of codebooks to keep for each super-pixel, $1\leq m\leq K$. Given the predicted weight map $W\!\in\!\mathbb{R}^{u\times v\times K}$, the encoder masks out $K-m$ items in each vector $\mathbf{w}_l\in\mathbb{R}^{K}$ corresponding to the $l$-th super-pixel ($l\!=\!1,\ldots,u\!\times\!v$). The masked out items have smallest absolute values to minimize the impact on the degraded latent $\Tilde{Y}$. Then for each super-pixel, only the non-zero remaining weights (16 bits per weight item) and the corresponding codebook indices (floor($\log_2K$) bits per weight item) need to be transmitted, totalling $(16+\text{floor}(\log_2K))\!\times\!m$ instead of the original $16\!\times\!K$. Parameter $m$ provides the tradeoff between bitrate and reconstruction quality. In general, the larger the number of codebooks to use, the better the reconstruction quality and the larger the bitrate. 

On the decoder side, using the received masked weight map $\Tilde{W}$, the degraded latent $\Tilde{Y}$ can be computed in the same way as Equation~(\ref{Eqn:weightedcombine}), where only the corresponding codebooks with non-zero weights contribute to the feature computation for each super-pixel. Based on this degraded latent $\Tilde{Y}$, the weight filler network predicts another full weight map $\hat{W}$ as a refilled version of the original weight map $W$. This refilled $\hat{W}$ is used to weighted combine quantized latent features $Y^q_1,\ldots,Y^q_K$ to recover the latent $\hat{Y}$, which is used to reconstruct the output image. 

Specifically, the weight filler has the same network structure with the weight predictor in \cite{AdaCode}, consisting of four residual swin transformer blocks (RSTBs) \cite{swinTransformer} and a convolution layer to match the channels of weight map and codebook number $K$. 

\subsection{Single Codebook Setup}
\label{sec:singlecodebook}

The above weight masking and refilling mechanism can be further optimized when only one codebook is used for each super-pixel. That is, we can further reduce the transmission bits by slightly modifying the weight predictor network, so that we do not need to transfer any weight parameters to the decoder. Specifically, a gumbel softmax layer \cite{gumbelsoftmax} is added onto the weight predictor so that one-hot weight entry is obtained for each super-pixel indicating the codebook to be used with importance weight as 1. In other words, only the  $u\times v\times \text{floor}(\log_2K)$ bits for codebook indices need to be transmitted to the decoder to retrieve the degraded latent $\Tilde{Y}$.  

It is worth mentioning that an intuitive alternative of the above single codebook setting is to treat all basis codebooks as one big codebook and skip weight prediction, where one codeword index is assigned to each super-pixel in the combined codebook. However, this alternative does not work in practice since the basis codebooks are learned separately, making it hard to directly compare their codeword features to obtain a cohesive index due to the scale difference. 

\subsection{Training Process}
\label{sec:training}

We adopt the embedding network $E^{emb}$ and the pre-trained semantic-class-dependent basis codebooks $\mathcal{C}_1,\ldots,\mathcal{C}_K$ from AdaCode \cite{AdaCode}, which partition the latent feature space into non-overlapping cells in $K$ different ways. They are kept fixed during our training process. Then we train the weight predictor, weight filler, the reconstruction network, and the GAN discriminator. On image level, the L1 loss $\mathcal{L}_1(\hat{x},x)$, the pereptual loss $\mathcal{L}_{per}(\hat{x},x)$ \cite{perceptualloss} and the adversarial loss $\mathcal{L}_{adv}(\hat{x},x)$ \cite{adversarialloss} are minimized to reduce the distortion between reconstructed $\hat{x}$ and input $x$. On feature level, the contrastive loss $\mathcal{L}_{con}(\hat{Y}, Y)$ \cite{SimCLR} is minimized to regularize the recovered latent $\hat{Y}$. Same as \cite{AdaCode}, the straight-through gradient estimator \cite{straightthrough} is used for back-propagating the non-differentiable vector quantization process during training. 

\begin{figure*}[t]
  \centering
   \includegraphics[width=\linewidth]{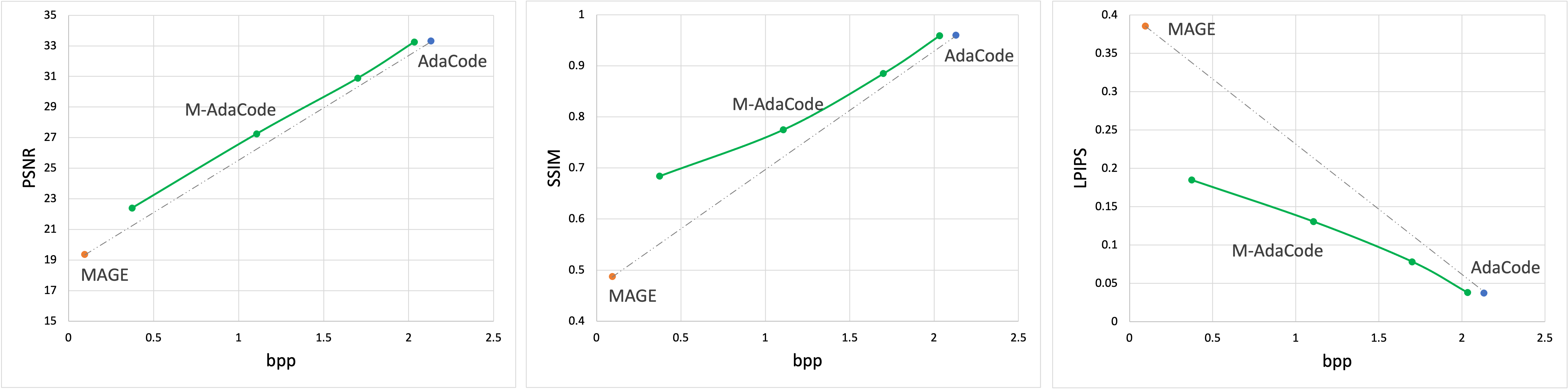}
   \caption{Quantitative comparison with SOTA SVR-based compression methods. \textbf{PSNR}/\textbf{SSIM}: the higher, the better. \textbf{LPIPS}: the lower, the better. Previous MAGE \cite{MAGE} and AdaCode \cite{AdaCode} operate with very low or high bitrates. M-Adacode provides better rate-distortion tradeoffs over a range of bitrates. }
   \label{fig:svr_performance}
\end{figure*}

\section{Experiments}

\noindent \textbf{Experimental Setup} Our experiments are based on the JPEG-AI dataset \cite{JPEG-AI,JPEG-AI-data}, which has 5664 images with a large variety of visual content and resolutions up to 8K. The training, validation, and test set have 5264, 350, and 50 images, respectively. The dataset is developed by the JPEG standardization organization to provide standard tools to evaluate NIC methods in the field. 

Following similar procedures as AdaCode \cite{AdaCode}, the training patches have $512\!\times\!512$ resolution, which are firstly randomly cropped from the training images, and then degraded by using the degradation model of BSRGAN \cite{BSRGAN}. For test evaluation, the maximum resolution of inference tiles is $1080\!\times\!1080$. 

The training stage has 200K iterations with Adam optimizer and a batch size of 64, using 8 NVIDIA Tesla V100 GPUs. The learning rate for the generator and discriminator are fixed as 1e-4 and 4e-4, respectively. \vspace{.5em}

\noindent \textbf{Evaluation Metrics} For reconstruction distortion, we measure PSNR and SSIM, as well as the perceptual LPIPS \cite{lpips}. The bitrate is measured by bpp (bit-per-pixel): $bpp\!=\!B/h\!\times\!w$. The overall bits $B\!=\!b_c\!+\!b_w$ consist of $b_{c}$ for transmitting codebook indices $Z_1\!\ldots,Z_K$ and $b_w$ for trainsmitting the sparse weight map $\Tilde{W}$. The naive calculation is $b_c\!=u\!\times\!v\!\times\!\sum_{k=1}^K\text{floor}(\log_2 n_k)$ ($n_k$ is the codebook size for $\mathcal{C}_k$). There are many methods to efficiently reduce $b_{c}$ by losslessly compressing the integer codebook indices, such as \cite{codecompress,codecompress2} with at lease $2\!\times$ to $3\!\times$ bit reduction. Reducing $b_c$ is a universal topic for SVR-based compression, which is out of the scope of this paper. We focus on reducing $b_w$ to trade off reconstruction quality for bitrate. For $b_w$, the required bits for each super-pixel falls into the range of $[\text{floor}(\log_2K),K\!\times\!16]$, where the minimum $\text{floor}(\log_2K)$ corresponds to the single-codebook setting discussed in Section.~\ref{sec:singlecodebook}, and the maximum $K\!\times\!16$ corresponds to AdaCode \cite{AdaCode}. For other cases using $m$ codebooks for each super-pixel ($1\!<\!m\!<\!K$), we have $b_w=u\!\times\!v\!\times(16+\text{floor}(\log_2K))\!\times\!m$.

\begin{figure*}[t]

    \centering
    \begin{minipage}[b]{0.225\linewidth}
    \centering
    \centerline{\small \textcolor{white}{g}Ground truth ($2500\!\times\!1667$)\textcolor{white}{g}}
    \vspace{0.02cm}
    \includegraphics[width=\textwidth]{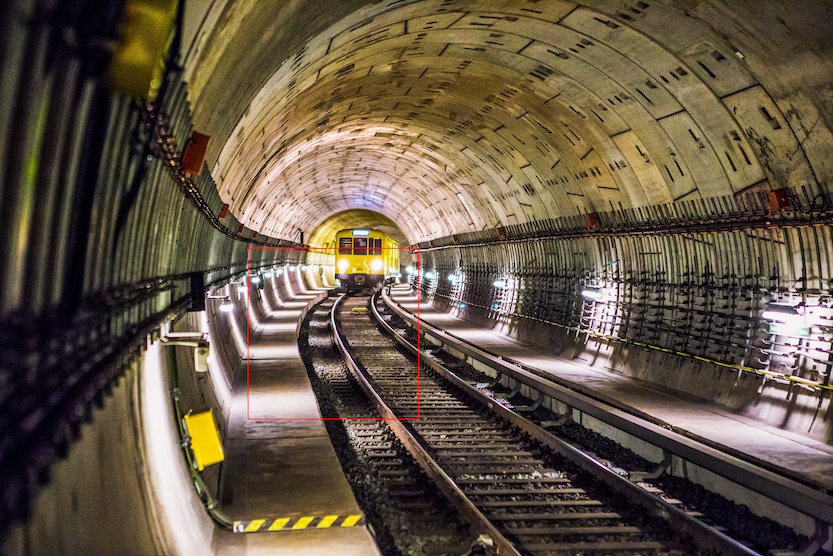}
    \vspace{-0.3cm}
    \centerline{\small\textcolor{white}{0.413$\mid$}}\medskip
    \end{minipage}
    \hspace{0.58cm}
    \begin{minipage}[b]{0.15\linewidth}
    \centering
    \centerline{\small \textcolor{white}{g}AdaCode\textcolor{white}{g}}
    \vspace{0.02cm}
    \includegraphics[width=\textwidth]{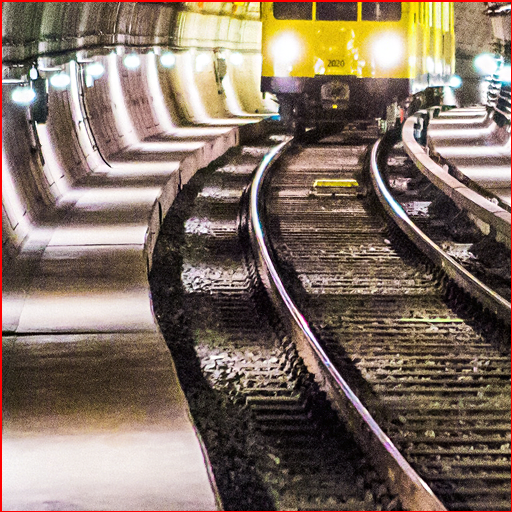}
    \vspace{-0.3cm}
    \centerline{\small 0.033$\mid$32.37$\mid$0.946}\medskip
    \end{minipage}
    \hspace{0.58cm}
    \begin{minipage}[b]{0.15\linewidth}
    \centering
    \centerline{\small \textcolor{white}{g}M-AdaCode 2-codebook\textcolor{white}{g}}
    \vspace{0.02cm}
    \includegraphics[width=\textwidth]{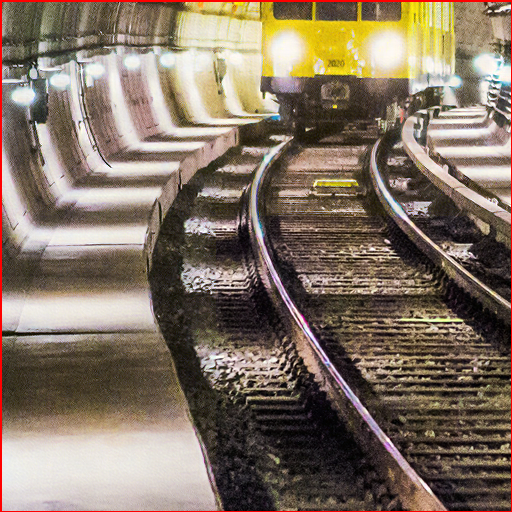}
    \vspace{-0.3cm}
    \centerline{\small0.153$\mid$25.68$\mid$0.754}\medskip
    \end{minipage}
    \hspace{0.58cm}
    \begin{minipage}[b]{0.15\linewidth}
    \centering
    \centerline{\small  \textcolor{white}{g}M-AdaCode 1-codebook\textcolor{white}{g}}
    \vspace{0.02cm}
    \includegraphics[width=\textwidth]{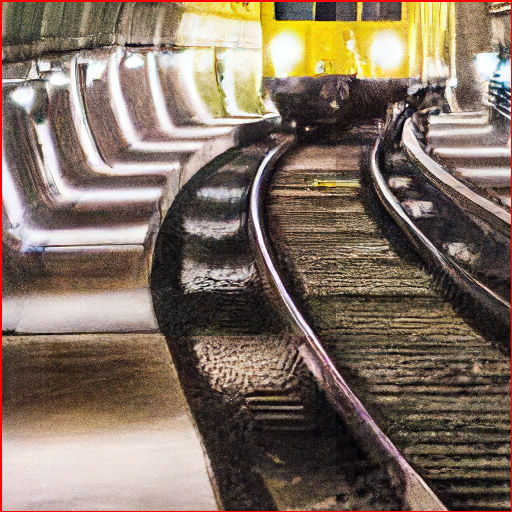}
    \vspace{-0.3cm}
    \centerline{\small0.184$\mid${21.37}$\mid$0.461}\medskip
    \end{minipage}
    \hspace{0.58cm}
    \begin{minipage}[b]{0.15\linewidth}
    \centering
    \centerline{\small \textcolor{white}{g}MAGE\textcolor{white}{g}}
    \vspace{0.02cm}
    \includegraphics[width=\textwidth]{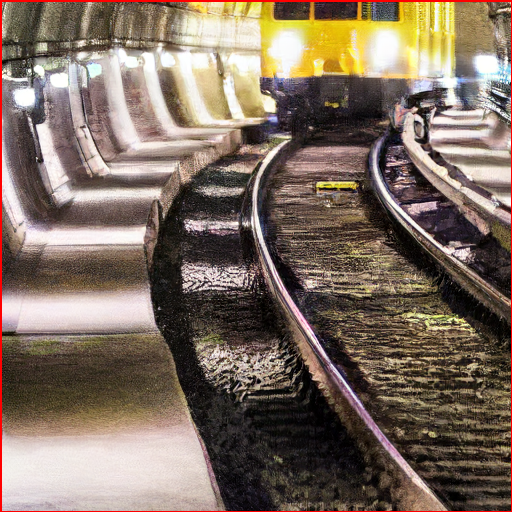}
    \vspace{-0.3cm}
    \centerline{\small{0.213}$\mid$18.91$\mid$0.289}\medskip
    \end{minipage}
    \vspace{.5em}

        \centering
    \begin{minipage}[b]{0.225\linewidth}
    \centering
    \centerline{\small \textcolor{white}{g}Ground truth ($2096\!\times\!1400$)\textcolor{white}{g}}
    \vspace{0.02cm}
    \includegraphics[width=\textwidth]{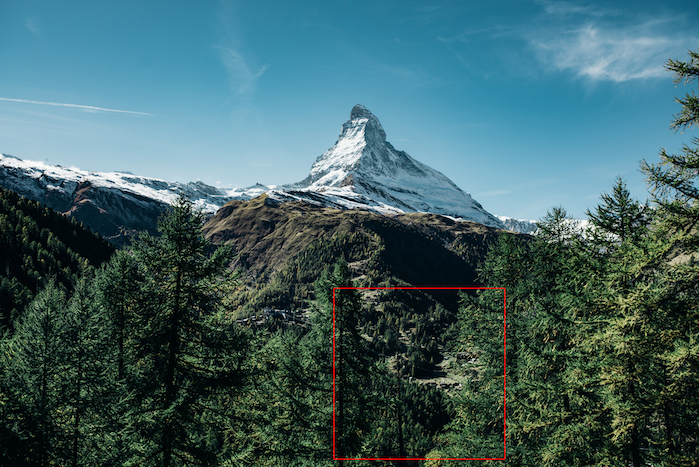}
    \vspace{-0.3cm}
    \centerline{\small\textcolor{white}{0.413$\mid$}}\medskip
    \end{minipage}
    \hspace{0.58cm}
    \begin{minipage}[b]{0.15\linewidth}
    \centering
    \centerline{\small \textcolor{white}{g}AdaCode\textcolor{white}{g}}
    \vspace{0.02cm}
    \includegraphics[width=\textwidth]{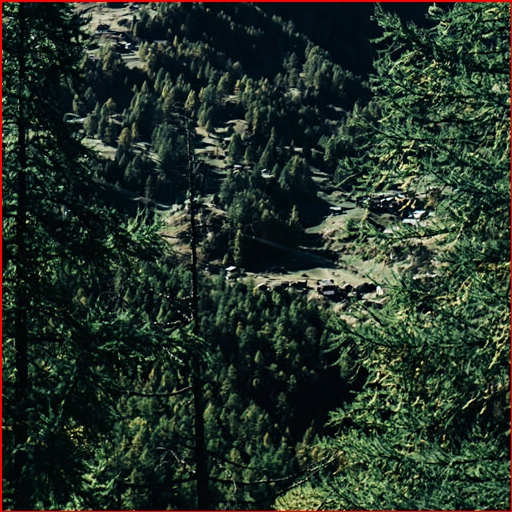}
    \vspace{-0.3cm}
    \centerline{\small 0.056$\mid$31.2$\mid$0.967}\medskip
    \end{minipage}
    \hspace{0.58cm}
    \begin{minipage}[b]{0.15\linewidth}
    \centering
    \centerline{\small \textcolor{white}{g}M-AdaCode 2-codebook\textcolor{white}{g}}
    \vspace{0.02cm}
    \includegraphics[width=\textwidth]{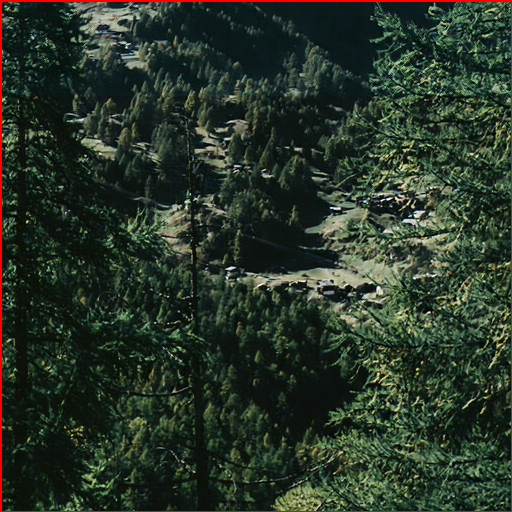}
    \vspace{-0.3cm}
    \centerline{\small 0.184$\mid$22.24$\mid$0.664}\medskip
    \end{minipage}
    \hspace{0.58cm}
    \begin{minipage}[b]{0.15\linewidth}
    \centering
    \centerline{\small  \textcolor{white}{g}M-AdaCode 1-codebook\textcolor{white}{g}}
    \vspace{0.02cm}
    \includegraphics[width=\textwidth]{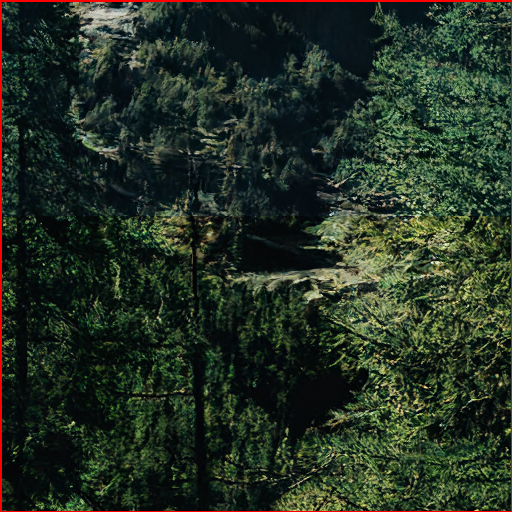}
    \vspace{-0.3cm}
    \centerline{\small 0.213$\mid${19.27}$\mid${0.602}}\medskip
    \end{minipage}
    \hspace{0.58cm}
    \begin{minipage}[b]{0.15\linewidth}
    \centering
    \centerline{\small \textcolor{white}{g}MAGE\textcolor{white}{g}}
    \vspace{0.02cm}
    \includegraphics[width=\textwidth]{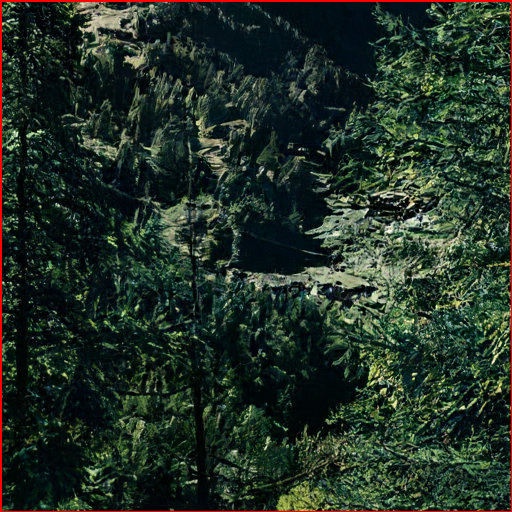}
    \vspace{-0.3cm}
    \centerline{\small0.295$\mid$17.22$\mid$0.476}\medskip
    \end{minipage}
    \vspace{.5em}

    \centering
    \begin{minipage}[b]{0.225\linewidth}
    \centering
    \centerline{\small \textcolor{white}{g}Ground truth ($1472\!\times\!976$)\textcolor{white}{g}}
    \vspace{0.02cm}
    \includegraphics[width=\textwidth,height=0.658\textwidth]{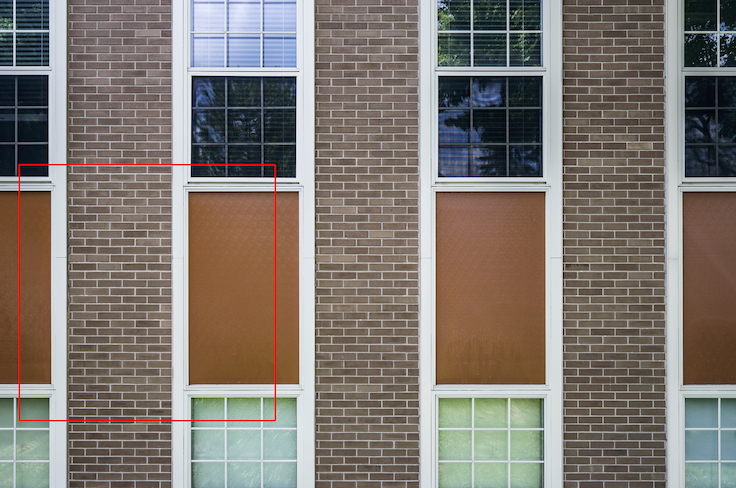}
    \vspace{-0.3cm}
    \centerline{\small\textcolor{white}{0.413$\mid$}}\medskip
    \end{minipage}
    \hspace{0.58cm}
    \begin{minipage}[b]{0.15\linewidth}
    \centering
    \centerline{\small \textcolor{white}{g}AdaCode\textcolor{white}{g}}
    \vspace{0.02cm}
    \includegraphics[width=\textwidth]{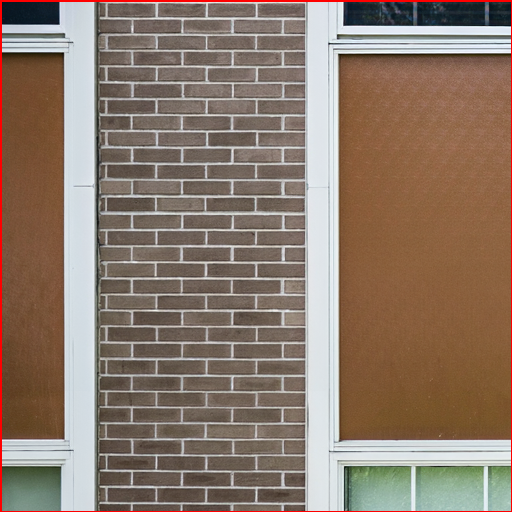}
    \vspace{-0.3cm}
    \centerline{\small 0.027$\mid$34.55$\mid$0.977}\medskip
    \end{minipage}
    \hspace{0.58cm}
    \begin{minipage}[b]{0.15\linewidth}
    \centering
    \centerline{\small \textcolor{white}{g}M-AdaCode 2-codebook\textcolor{white}{g}}
    \vspace{0.02cm}
    \includegraphics[width=\textwidth]{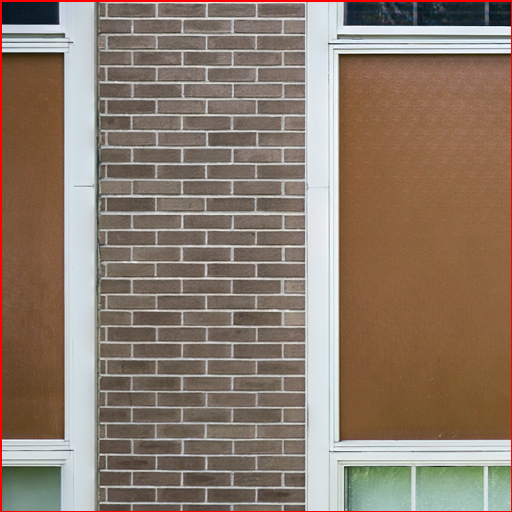}
    \vspace{-0.3cm}
    \centerline{\small 0.086$\mid$27.62$\mid$0.846}\medskip
    \end{minipage}
    \hspace{0.58cm}
    \begin{minipage}[b]{0.15\linewidth}
    \centering
    \centerline{\small  \textcolor{white}{g}M-AdaCode 1-codebook\textcolor{white}{g}}
    \vspace{0.02cm}
    \includegraphics[width=\textwidth]{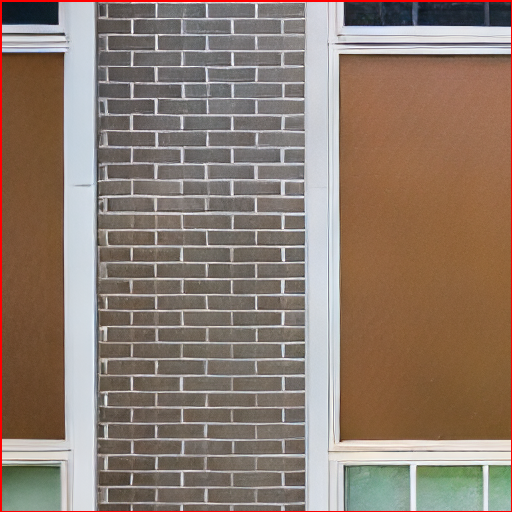}
    \vspace{-0.3cm}
    \centerline{\small 0.116$\mid${22.54}$\mid${0.769}}\medskip
    \end{minipage}
    \hspace{0.58cm}
    \begin{minipage}[b]{0.15\linewidth}
    \centering
    \centerline{\small \textcolor{white}{g}MAGE\textcolor{white}{g}}
    \vspace{0.02cm}
    \includegraphics[width=\textwidth]{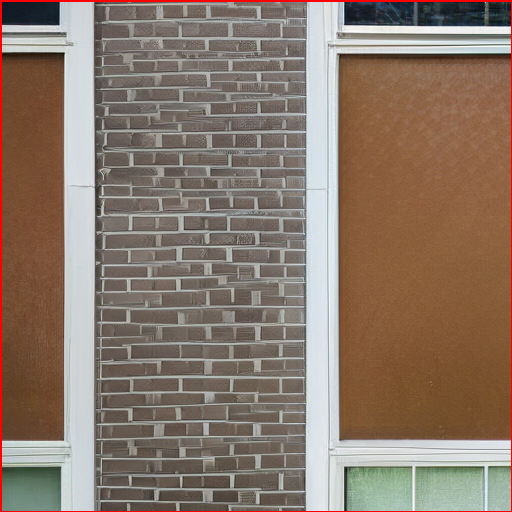}
    \vspace{-0.3cm}
    \centerline{\small{0.290}$\mid$19.34$\mid$0.461}\medskip
    \end{minipage}
    \vspace{.5em}
    
    \centering
    \begin{minipage}[b]{0.225\linewidth}
    \centering
    \centerline{\small \textcolor{white}{g}Ground truth ($1192\!\times\!832$)\textcolor{white}{g}}
    \vspace{0.02cm}
    \includegraphics[width=\textwidth,height=0.658\textwidth]{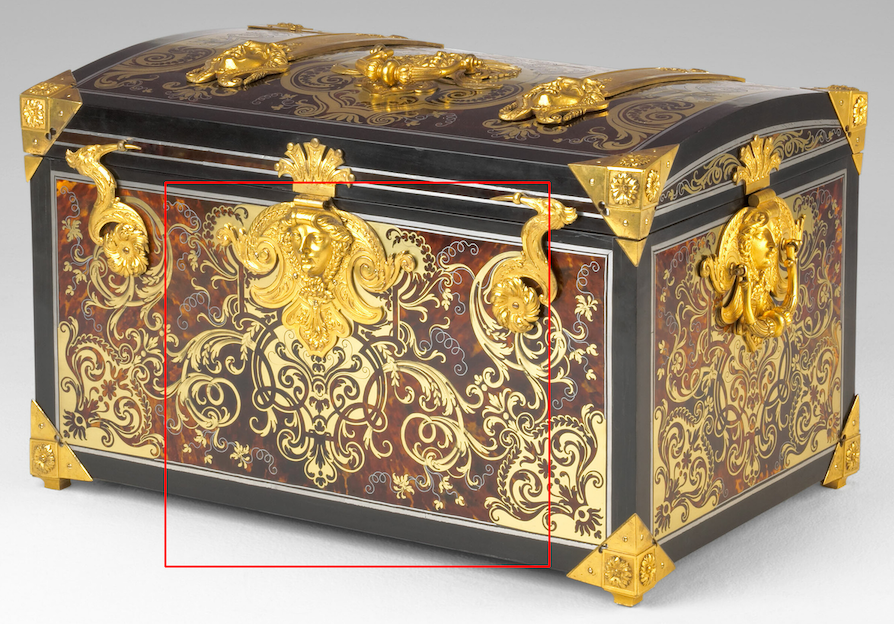}
    \vspace{-0.3cm}
    \centerline{\small\textcolor{white}{0.413$\mid$}}\medskip
    \end{minipage}
    \hspace{0.58cm}
    \begin{minipage}[b]{0.15\linewidth}
    \centering
    \centerline{\small \textcolor{white}{g}AdaCode\textcolor{white}{g}}
    \vspace{0.02cm}
    \includegraphics[width=\textwidth]{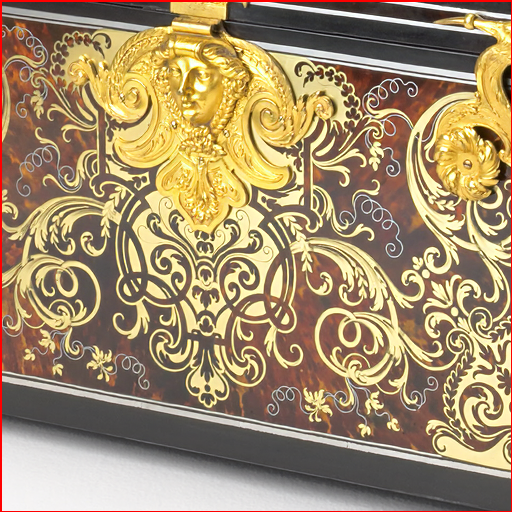}
    \vspace{-0.3cm}
    \centerline{\small0.028$\mid$31.01$\mid$0.971}\medskip
    \end{minipage}
    \hspace{0.58cm}
    \begin{minipage}[b]{0.15\linewidth}
    \centering
    \centerline{\small \textcolor{white}{g}M-AdaCode 2-codebook\textcolor{white}{g}}
    \vspace{0.02cm}
    \includegraphics[width=\textwidth]{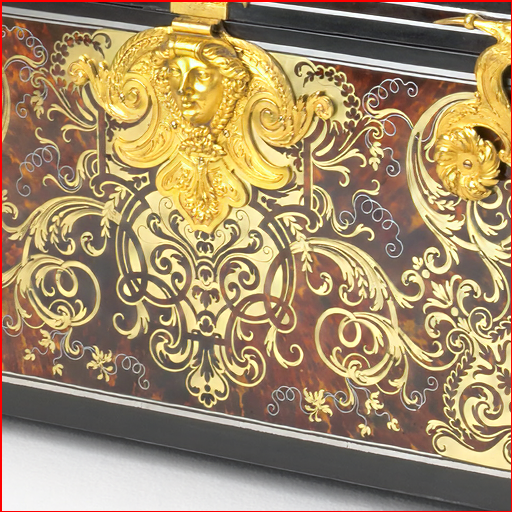}
    \vspace{-0.3cm}
    \centerline{\small0.039$\mid$27.97$\mid$0.766}\medskip
    \end{minipage}
    \hspace{0.58cm}
    \begin{minipage}[b]{0.15\linewidth}
    \centering
    \centerline{\small \textcolor{white}{g}M-AdaCode 1-codebook\textcolor{white}{g}}
    \vspace{0.02cm}
    \includegraphics[width=\textwidth]{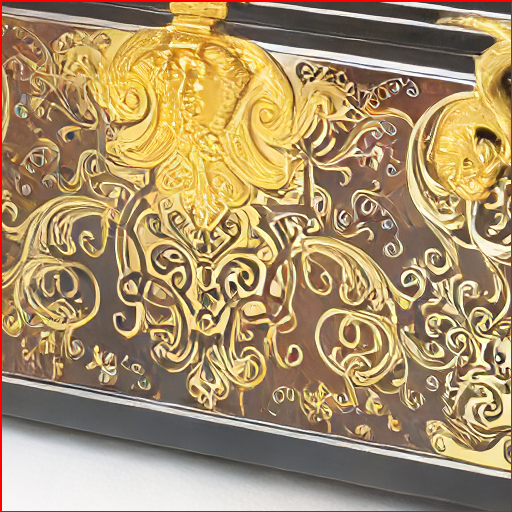}
    \vspace{-0.3cm}
    \centerline{\small 0.046$\mid${20.07}$\mid${0.688}}\medskip
    \end{minipage}
    \hspace{0.58cm}
    \begin{minipage}[b]{0.15\linewidth}
    \centering
    \centerline{\small \textcolor{white}{g}MAGE\textcolor{white}{g}}
    \vspace{0.02cm}
    \includegraphics[width=\textwidth]{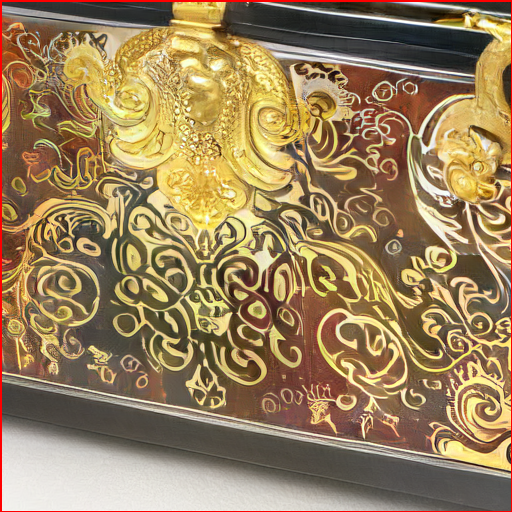}
    \vspace{-0.3cm}
    \centerline{\small{0.051}$\mid$16.04$\mid$0.473}\medskip
    \end{minipage}
    \vspace{.5em}

    \centering
    \begin{minipage}[b]{0.225\linewidth}
    \centering
    \centerline{\small \textcolor{white}{g}Ground truth ($1200\!\times\!800$)\textcolor{white}{g}}
    \vspace{0.02cm}
    \includegraphics[width=\textwidth,height=0.658\textwidth]{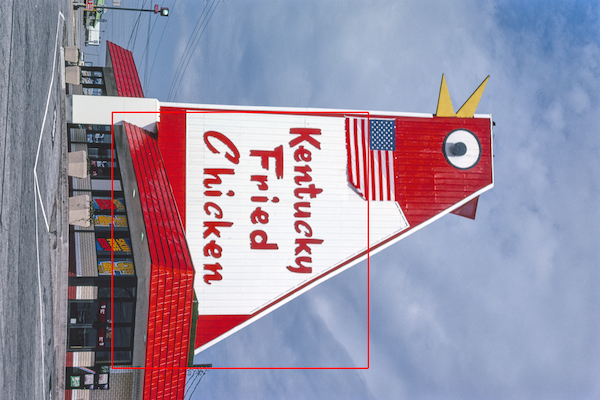}
    \vspace{-0.3cm}
    \centerline{\small\textcolor{white}{0.413$\mid$}}\medskip
    \end{minipage}
    \hspace{0.58cm}
    \begin{minipage}[b]{0.15\linewidth}
    \centering
    \centerline{\small \textcolor{white}{g}AdaCode\textcolor{white}{g}}
    \vspace{0.02cm}
    \includegraphics[width=\textwidth]{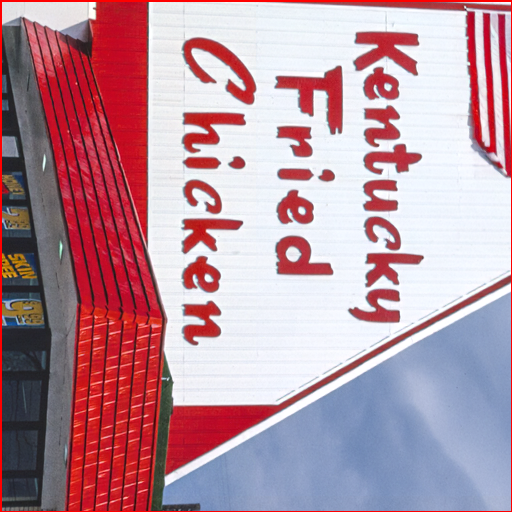}
    \vspace{-0.3cm}
    \centerline{\small 0.031$\mid$35.30$\mid$0.972}\medskip
    \end{minipage}
    \hspace{0.58cm}
    \begin{minipage}[b]{0.15\linewidth}
    \centering
    \centerline{\small \textcolor{white}{g}M-AdaCode 2-codebook\textcolor{white}{g}}
    \vspace{0.02cm}
    \includegraphics[width=\textwidth]{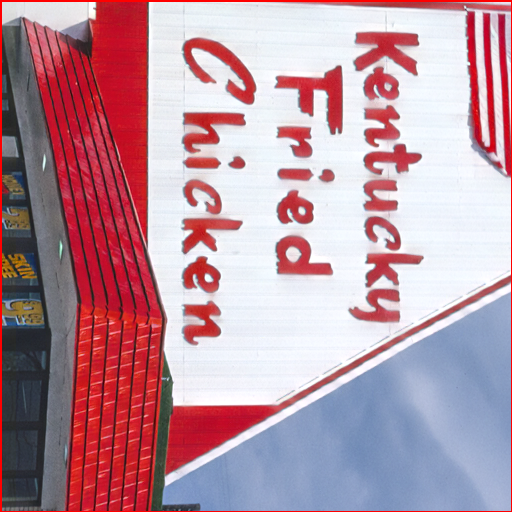}
    \vspace{-0.3cm}
    \centerline{\small 0.094$\mid$28.01$\mid$0.882}\medskip
    \end{minipage}
    \hspace{0.58cm}
    \begin{minipage}[b]{0.15\linewidth}
    \centering
    \centerline{\small  \textcolor{white}{g}M-AdaCode 1-codebook\textcolor{white}{g}}
    \vspace{0.02cm}
    \includegraphics[width=\textwidth]{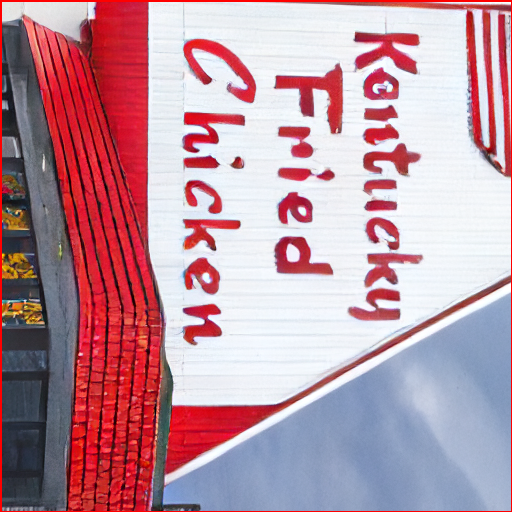}
    \vspace{-0.3cm}
    \centerline{\small 0.113$\mid${25.04}$\mid${0.827}}\medskip
    \end{minipage}
    \hspace{0.58cm}
    \begin{minipage}[b]{0.15\linewidth}
    \centering
    \centerline{\small \textcolor{white}{g}MAGE\textcolor{white}{g}}
    \vspace{0.02cm}
    \includegraphics[width=\textwidth]{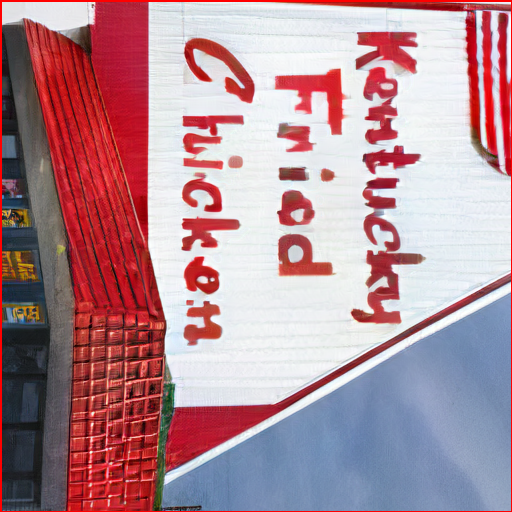}
    \vspace{-0.3cm}
    \centerline{\small{0.145}$\mid$19.74$\mid$0.647}\medskip
    \end{minipage}
    \vspace{.5em}

    \centering
    \begin{minipage}[b]{0.225\linewidth}
    \centering
    \centerline{\small \textcolor{white}{g}Ground truth ($945\!\times\!840$)\textcolor{white}{g}}
    \vspace{0.02cm}
    \includegraphics[width=\textwidth,height=0.658\textwidth]{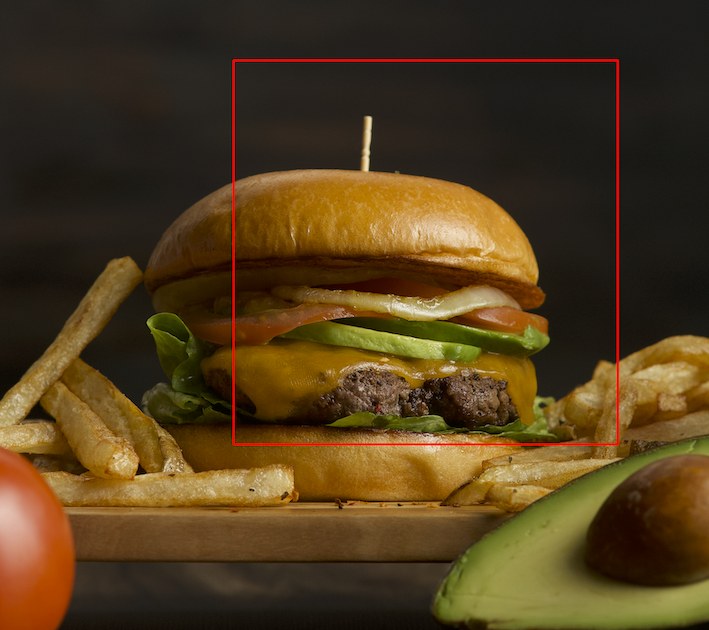}
    \vspace{-0.3cm}
    \centerline{\small\textcolor{white}{0.413$\mid$}}\medskip
    \end{minipage}
    \hspace{0.58cm}
    \begin{minipage}[b]{0.15\linewidth}
    \centering
    \centerline{\small \textcolor{white}{g}AdaCode\textcolor{white}{g}}
    \vspace{0.02cm}
    \includegraphics[width=\textwidth]{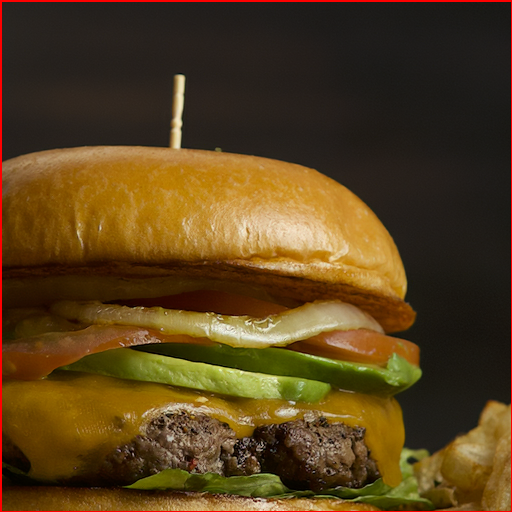}
    \vspace{-0.3cm}
    \centerline{\small 0.025$\mid$39.98$\mid$0.979}\medskip
    \end{minipage}
    \hspace{0.58cm}
    \begin{minipage}[b]{0.15\linewidth}
    \centering
    \centerline{\small \textcolor{white}{g}M-AdaCode 2-codebook\textcolor{white}{g}}
    \vspace{0.02cm}
    \includegraphics[width=\textwidth]{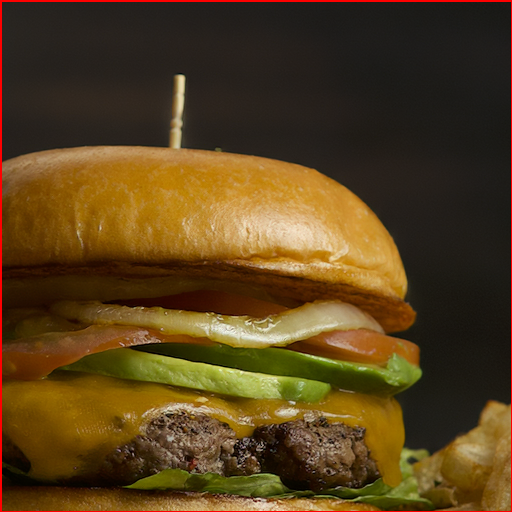}
    \vspace{-0.3cm}
    \centerline{\small0.060$\mid$34.99$\mid$0.906}\medskip
    \end{minipage}
    \hspace{0.58cm}
    \begin{minipage}[b]{0.15\linewidth}
    \centering
    \centerline{\small  \textcolor{white}{g}M-AdaCode 1-codebook\textcolor{white}{g}}
    \vspace{0.02cm}
    \includegraphics[width=\textwidth]{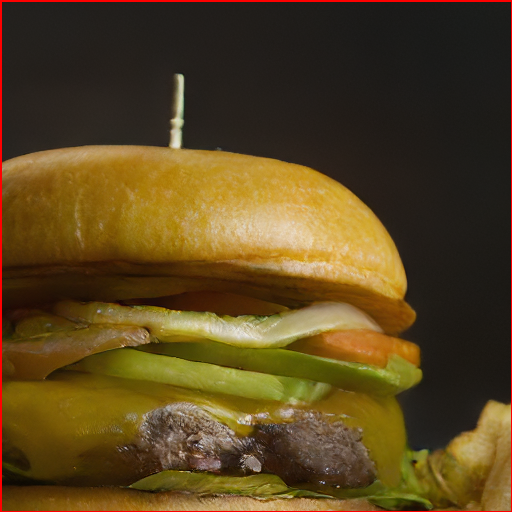}
    \vspace{-0.3cm}
    \centerline{\small 0.108$\mid${27.23}$\mid${0.883}}\medskip
    \end{minipage}
    \hspace{0.58cm}
    \begin{minipage}[b]{0.15\linewidth}
    \centering
    \centerline{\small \textcolor{white}{g}MAGE\textcolor{white}{g}}
    \vspace{0.02cm}
    \includegraphics[width=\textwidth]{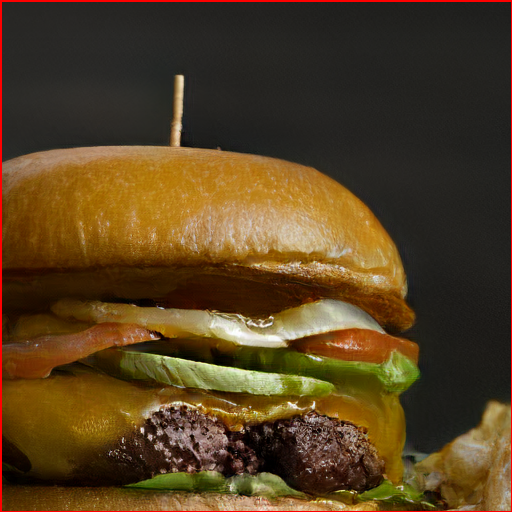}
    \vspace{-0.3cm}
    \centerline{\small{0.132}$\mid$23.56$\mid$0.746}\medskip
    \end{minipage}
    \vspace{-.5em}

\caption{Reconstruction examples. Numbers under each result are ``LPIPS$|$PSNR$|$SSIM". ``M-AdaCode 1-codebook'' and ``M-AdaCode 2-codebook'' are M-AdaCode using 1 codebook or 2 codebooks per super-pixel, respectively.}
\label{fig:examples}
\end{figure*}

\subsection{Reconstruction performance}

Figure~\ref{fig:svr_performance} gives the rate-distortion comparison of different methods. For M-AdaCode, the performance under 4 settings are tested, where each super-pixel uses $m=1,\ldots,4$ codebooks, respectively. The bit counts $b_c$ shown in the figure are computed by simply using the zip software to compress the integer codebook indices, which gives roughly $2\!\times$ bit reduction comparing to the naive calculation. From the figure, MAGE and AdaCode operate as SVR-based compression methods for extreme scenarios. MAGE targets at a very low bitrate ($<0.1$ bpp) with perceptually reasonable generation. AdaCode targets at high reconstruction quality but has a very high bitrate ($>2$ bpp). The dotted line connecting these two methods are the conceptual rate-distortion tradeoffs that an SVR-based compression method should be able to provide based on previous methods. As shown in the figure, our M-AdaCode can operate over a wide range of bitrates in between, and can give much better rate-distortion tradeoffs. Table~\ref{tab:gains} summarizes the performance gains M-AdaCode achieves comparing to the conceptual baseline. Basically, M-Adacode performs much better in terms of SSIM and perceptual LPIPS. The improvements over PSNR are not as significant. This is as expected since the strength of generative methods is to generate rich details to improve perceptual quality, and such rich details do not necessarily match original inputs at the pixel level. 

\begin{table}
  \centering
  \begin{tabular}{|c|c|c|c|}
    \hline
    bpp & PSNR & SSIM & LPIPS\\\hline\hline
    0.373 & 5.3\% & 23.8\% & 45.3\%\\\hline
    1.016 & 3.6\% & 7.2\% & 60.4\%\\\hline
    1.701 & 1.7\% & 2.8\% & 29.5\%\\\hline
    2.033 & 1.8\% & 2.3\% & 29.8\%\\
    \hline
  \end{tabular}
  \caption{Improvements of M-AdaCode over the conceptual baseline. }
  \label{tab:gains}
\end{table}

Figure~\ref{fig:examples} gives some examples of the reconstruction results comparing different methods, for images with different visual content and with different resolutions. The corresponding quantitative performance of these examples are also listed. As clearly shown in the figure, by transferring the full weight map, ``AdaCode" can recover rich and accurate details. Using a single codebook per super-pixel, ``M-AdaCode 1-codebook'' can generate visually pleasing results with reasonable details while preserving good fidelity to the ground-truth. In comparison, using one generic codebook without image-adaptive information, the reconstructed image using ``MAGE" presents lots of artifacts or inconsistent details. In many cases, using only two codebooks per super-pixel, ``M-AdaCode 2-codebook" can reconstruct images with quite good visual quality. 

\begin{figure*}[t]
  \centering
   \includegraphics[width=\linewidth]{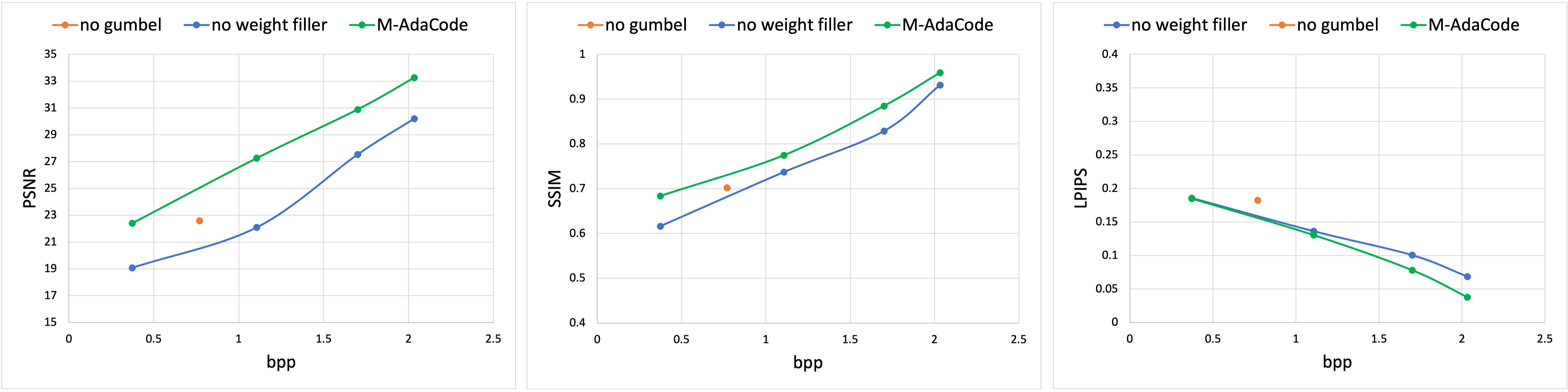}
   \caption{Ablation study: performance without weight filler and performance without single codebook setting. Weight filler can largely improve pixel-level distortion, and the single codebook setting can reduce bitrate without hurting distortion.}
   \label{fig:ablation}
\end{figure*}

\subsection{Ablation Study}

In this section, we investigate the importance of weight filler and the effectiveness of the single codebook setting of Section \ref{sec:singlecodebook}.  Without the weight filler, the decoder directly uses degraded latent $\Tilde{Y}$ to reconstruct output $\hat{x}$. In this case, only the weight predictor and reconstruction network are trained in the training process of Section \ref{sec:training}. Without the single codebook setting, the same network structure (without gumbel softmax) for the weight predictor is used when only one codebook is kept for each super-pixel, and the bit count for the weight map is $b_w\!=\!u\!\times\!v\!\times\!(16\!+\!\text{floor}(\log_2K))$.

Figure~\ref{fig:ablation} gives the performance comparison with M-AdaCode. When one codebook is used for each super-pixel, the single codebook setting can achieve equivalent distortion performance with a 52\% reduction on bitrate. Using the weight filler, the performance of pixel-level PSNR and SSIM are significantly better than direct reconstruction from the degraded latent feature, especially for lower bitrates. The influence of weight filler reduces as the bitrate increases. As for LPIPS, even without weight filler, by training good reconstruction network the generated image still has reasonable perceptual quality.

\subsection{More Discussions}

\noindent\textbf{Advantages} As mentioned before, SVR-based compression has the advantage of being robust against small transmission and calculation errors across heterogeneous hardware and software platforms. Moreover, the proposed M-AdaCode framework has some additional appealing features. 

First, the granularity of the learened basis codebooks to model the separated latent space impacts the reconstruction quality. In general, more basis codebooks with finer granularity give better reconstruction quality, but with a price of larger bitrates. M-AdaCode gives a method to trade off distortion and bitrate. Potentially, we can pretrain many basis codebooks to model the vast visual content space, and customize a limited number of codebooks for each particular data domain based on practical needs.  

Second, the dimensionality of the latent feature space, \textit{i.e.}, the codeword feature dimension, also impacts the reconstruction quality. Usually, more dimensions give more representation capacity, leading to better reconstruction but with the price of more storage and computation costs. When the codeword feature dimension increases, M-AdaCode does not increase the bitrate by transferring codeword indices. So potentially, we can use rich representation with large feature dimensions, as long as being permitted by the computation and storage requirements. 

\noindent\textbf{Limitations} As a generative image modeling method, the SVR-based compression has a competing goal of generative visual quality and pixel-level fidelity to the input. This is an advantage when the input has low or mediocre quality, especially when the input has degradations. In such cases, the target can be interpreted as to restore the conceptual high-quality clean input from the degraded version, and using high-quality codewords is robust to recover good visual details. However, when the input has ultra-high quality, the generated details may be inconsistent to the input and may hurt the performance, since in such cases the target is to recover the exact input itself. Therefore, in practical usage, it may be hard for a particular method to work universally better than others, and we may need to selectively choose which method to use when compressing images with different quality and different content.

\section{Conclusion}

We propose an SVR-based image compression method, M-AdaCode, by using masks over the latent feature subspace to balance bitrate and reconstruction quality. The encoder embeds images into discrete latent subspaces spanned by multiple basis codebooks that are learned in a semantic-class-dependent fashion, and transfers integer codeword indices that are efficient and cross-platform robust. By deriving image-adaptive weights to combine the basis codebooks, a rich latent feature can be recovered for high quality reconstruction. Using the redundant information in the latent subspaces, unimportant weights can be masked out in the encoder and recovered later in the decoder, to trade off reconstruction quality for transmission bits. The masking rate controls the balance between bitrate and distortion. Experiments over the standard JPEG-AI dataset show that comparing to previous SVR-based compression methods that operate over very low or very high bitrates, our M-AdaCode achieves better rate-distoration tradeoffs over a large range of bitrates.

{\small
\bibliographystyle{ieee_fullname}
\bibliography{egbib}
}

\end{document}